\documentclass[journal]{IEEEtran}
\usepackage{amsmath}

\usepackage[numbers, sort]{natbib}

\ifCLASSINFOpdf
   \usepackage[pdftex]{graphicx}
\else
\fi
\begin{document}
%
\title{Traffic Flow Forecasting with Maintenance Downtime via Multi-Channel Attention-Based Spatio-Temporal Graph Convolutional Networks}
%
%
%
\author{Yuanjie Lu\thanks{Department of Computer Science, George Mason University, Fairfax, VA, 22030. Email: ylu22@gmu.edu},
\and Parastoo Kamranfar\thanks{Department of Computer Science, George Mason University, Fairfax, VA, 22030. Email: pkamranf@gmu.edu},
\and David Lattanzi\thanks{Department of Civil Engineering, George Mason University, Fairfax, VA, 22030. Email: dlattanz@gmu.edu},
\and Amarda Shehu\thanks{Department of Computer Science, George Mason University, Fairfax, VA, 22030. Email: amarda@gmu.edu}

}
\maketitle


\begin{abstract} 
Forecasting traffic flows is a central task in intelligent transportation system management. Graph structures have shown promise as a modeling framework, with recent advances in spatio-temporal modeling via graph convolution neural networks, improving the performance or extending the prediction horizon on traffic flows. However, a key shortcoming of state-of-the-art methods is their inability to take into account information of various modalities, for instance the impact of maintenance downtime on traffic flows. This is the issue we address in this paper. Specifically, we propose a novel model to predict traffic speed under the impact of construction work. The model is based on the powerful attention-based spatio-temporal graph convolution architecture but utilizes various channels to integrate different sources of information, explicitly builds spatio-temporal dependencies among traffic states, captures the relationships between heterogeneous roadway networks, and then predicts changes in traffic flow resulting from maintenance downtime events. The model is evaluated on two benchmark datasets and a novel dataset we have collected over the bustling Tyson's corner region in Northern Virginia. Extensive comparative experiments and ablation studies show that the proposed model can capture complex and nonlinear spatio-temporal relationships across a transportation corridor, outperforming baseline models.
\end{abstract}

\section{Introduction}

\IEEEPARstart{G}{rowing} sophistication in deep learning is renewing attention in feature-free intelligent transportation system modeling for traffic management~\cite{Zhang11}. Advances in spatio-temporal modeling and neural network architectures that can handle graph data have lead to many architectures that are increasingly improving performance or extending the prediction horizon on traffic flow~\cite{cui2019traffic, li2017diffusion, yu2017spatio, diao2019dynamic}. 

Two main architectures have been popular in recent years. Convolutional neural networks (CNN) have been employed to extract spatial features of grid-based data and handle high-dimensional spatio-temporal data. Graph convolutional neural networks (GCN) have been shown to be more powerful due to their ability to describe spatial correlations of graph-based data. Both CNN- and GCN-based models cannot simultaneously model the spatio-temporal features and dynamic correlations of traffic data. A spatial-temporal attention mechanism is added in~\cite{guo2019attention} to learn the dynamic spatial-temporal correlations of traffic data; spatial attention models the complex spatial correlations between different locations, and a temporal attention captures the dynamic temporal correlations between different times.

A key shortcoming of current state-of-the-art (SOTA) methods for traffic flow forecasting is the inability to take into account information of different modalities. Predicting corridor traffic flow performance is a complex and difficult task with a high degree of statistical variance under most formulation. Most conventional traffic speed prediction methods are based on models that often cannot accommodate downtime and maintenance factors with reasonable granularity. Predictive performance also tends to break down as the scale and complexity of a transportation network increases. Additionally, most existing methods struggle to incorporate traffic factors such as road maintenance and restoration activities or traffic accidents. 

Downtime not only causes road congestion but can also create significant safety hazards. According to an urban mobility report released in 2019, the economic toll of traffic congestion has increased by nearly 48\% over the past ten years~\cite{schrank_2019_2019,du2017predicting}. Modeling and forecasting the impacts of maintenance downtime on a transportation corridor can provide engineers and managers with tools for minimizing disruptions and optimizing the logistics of maintenance, while also maintaining optimal traffic flow for the traveling public.

In this paper we build over the SOTA graph convolutional network framework and propose a novel GCN-based model for traffic flow forecasting under the additional effect of construction downtime incidents. We refer to the model as \underline{G}raph \underline{C}onvolutional \underline{N}etwork for \underline{R}oadway \underline{W}ork \underline{Z}ones (GCN-RWZ). The main contributions of this paper are: 

(1) We develop a multi-channel fusion approach to absorb multiple characteristics and handle various continuous values in predictive modeling. Descriptors of the various traffic network’s characteristics are included through a flexible “feature map” data format. GCN-RWZ is designed to capture the influence of construction impacts within a corridor of arbitrary scale, making it flexible and generalizable to a variety of corridors and regional conditions. (2) We construct a new dataset to evaluate the model and serve as a benchmark; it is currently difficult to find a dataset that contain additional information on construction workzones, so we build the dataset that contains all speed and corresponding information during each construction work. The dataset is collected over the busy Tyson’s Corner region of Northern Virginia. (3) Extensive comparative experiments and ablation studies on three real-world traffic datasets show that the proposed model can integrate diverse sources of information and capture complex and nonlinear spatio-temporal relationships across a transportation corridor, outperforming existing baseline models.

The rest of the paper is organized as follows. In Section~\ref{sec:RelatedWork} we provide a focused review of related work on state-of-the-art GCN-based architectures for traffic flow forecasting. The proposed method is then described in detail in Section~\ref{sec:Methods} and evaluated in Section~\ref{sec:Experiments}. The paper concludes with a summary of future work in Section~\ref{sec:Conclusions}. 
\section{Related Work}
\label{sec:RelatedWork}
Shallow machine learning models have been widely used to predict city-scale traffic flow, for instance ARIMA models~\cite{tong2008highway}, support vector regression~\cite{wu2004travel}, hybrid ensemble models including ANNs and bagging~\cite{moretti2015urban}, and spatial auto-regressive (SAR) models \cite{kelejian1999generalized}. Due to their abilities to incorporate disparate data types through expanded dimensionality, and to handle nonlinear data associations, these models can capture the spatial and temporal correlation in traffic data. However, their reliance on expert-crafted features is a key limitation that hampers their performance, generalizability, and adoption.

In response, researchers have turned to deep learning models that learn directly from data. Early work in traffic prediction has focused on deep belief networks~\cite{huang2014deep}, recurrent neural network (RNN)~\cite{tian2015predicting}, and long short-term memory (LSTM)~\cite{ma2015long} models that can model temporal correlations. More recently, researchers have used multi-model patterns to consider both temporal and spatial dependencies. For instance, work in~\cite{ma2015large} utilizes a deep Restricted Boltzmann Machine within an RNN architecture to capture features of traffic congestion. Work in~\cite{cui2018deep} proposes a deep bidirectional and unidirectional LSTM framework to measure backward dependencies. Recently, many researchers have been inspired by the capability of CNN-based frameworks (in the computer vision domain) to extract structured features and so have utilized CNN-based model to capture spatial correlations between traffic sensors. Work~\cite{ma2017learning} proposes a CNN-based method that models traffic as a set of large-scale, network-wide images. Subsequently, work in~\cite{jo2018image} develops a CNN to convert traffic states into an enhanced physical map. While these methods have improved prediction accuracy, they do not easily capture spatial relationships across a transportation corridor.

Graph neural networks (GNNs), a recent advancement in deep learning, can capture spatial correlations and are now popular in natural language processing (NLP), image, and speech recognition~\cite{vaswani2017attention}. Their utilization in transportation engineering is emerging~\cite{wu2019graph, guo2019attention, pan2019urban, ruiz2020gated, zheng2020gman, keneshloo2019deep, velivckovic2017graph}. A GNN can incorporate the topology of a road-level traffic network via the concept of a graph and so capture both spatial and temporal correlations. Some  GNN-based transportation research utilizes graph network embedding~\cite{perozzi2014deepwalk, kang2019learning, zheng2020gman} and recurrent graph neural networks (RecGNNs)~\cite{wu2020comprehensive, ruiz2020gated}. In addition to being computationally costly, these methods only transmit the information of each node and update the state of its own node, which cannot capture spatial relationships in a traffic network. To address this, current state-of-the-art methods use a particular network variant, the Graph Convolutional Network (GCN). Instead of iterating over states and propagating information from a sequence of nodes, GCNs attempt to support a graph with a fixed structure and build convolutional layers to extract the essential features. Such a model, pioneered in~\cite{zhao2019t}, predicts traffic speeds by combining GCN and the Gated Recurrent Unit (GRU) model; the GCN is used to learn topological structures for capturing the spatial correlations, and the GRU is used to learn variations of each tensor for capturing the temporal dependencies. In~\cite{zhou2020reinforced}, a new policy gradient is also proposed for updating the model parameters while alleviating bias.

GCN-based methods have now shown their power in traffic forecasting, but most studies mainly use the speed attribute for prediction. It should be noted that predictive performance depends on how much traffic speed is affected by anomalous factors, particularly the traffic speed under construction works; thus state-of-the-art methods currently cannot capture the spatio-temporal correlations of more complex traffic networks. Therefore, this paper investigates modeling the effects of maintenance downtime on a transportation corridor using deep neural networks designed to work with graph-structured data for the purpose of time-series traffic flow prediction.
\section{GCN-RWZ: Architecture and Methodology}
\label{sec:Methods}

\begin{figure*}[htbp]
\centering
\begin{tabular}{c}
\includegraphics[width=0.8\textwidth]{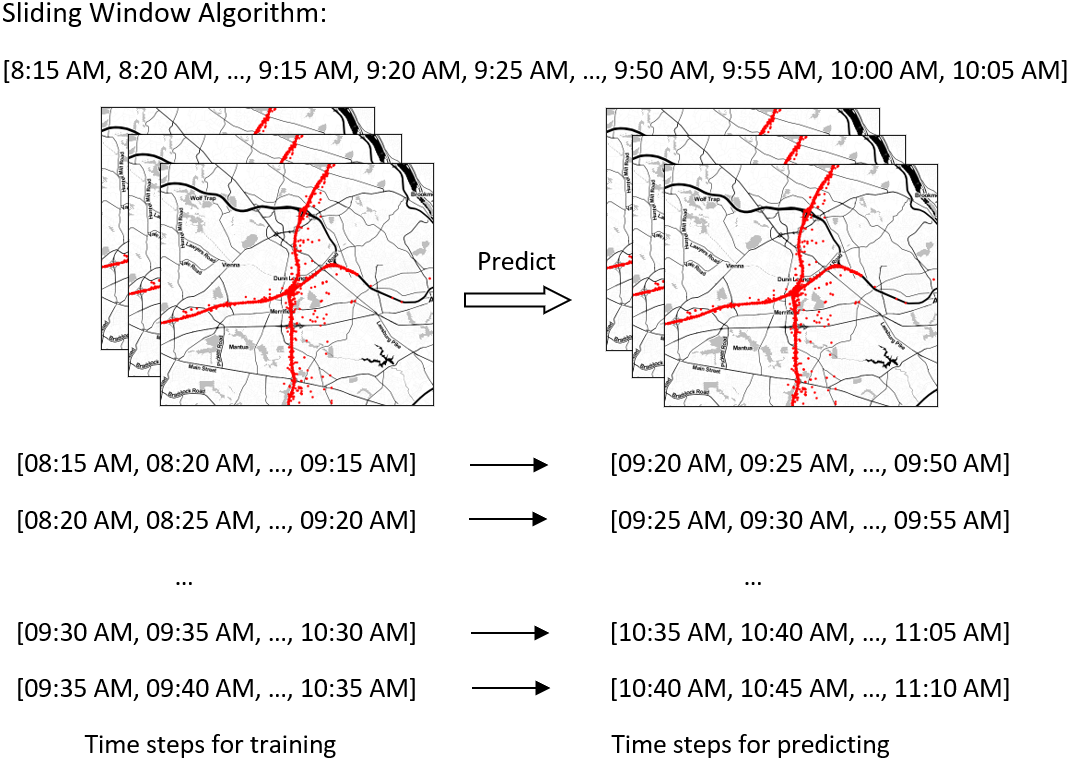}\\

\includegraphics[width=\textwidth]{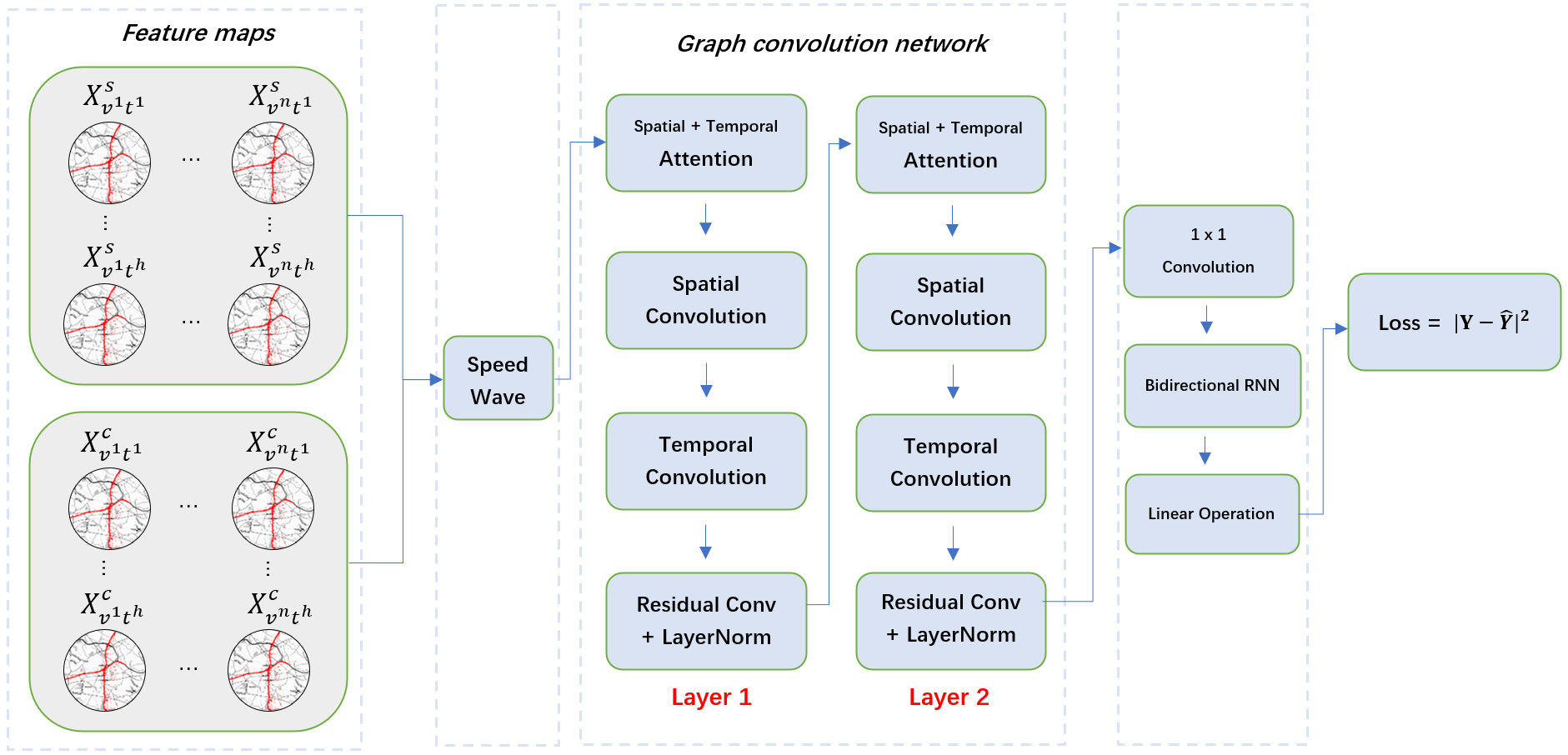}\\
\end{tabular}
\caption{Top: Traffic speed feature map time series, generated via a sliding window algorithm. Bottom: Overall framework of the developed flow prediction methodology and of the proposed GCN-RWZ model.}
\label{fig:InputAndFramework}
\end{figure*}

As summarized conceptually in Figure~\ref{fig:InputAndFramework}, the proposed GCN-RWZ model is capable of ingesting a variety of different feature maps associated with a range of network descriptors; time series data representing the flow of traffic in a corridor are fused with corresponding information on construction work. The top panel in Figure~\ref{fig:InputAndFramework} shows the traffic speed feature map time series generated via a window sliding algorithm we describe in greater detail below. The combined traffic and construction data representation is referred to here as a ``speed wave.'' The bottom panel of Figure~\ref{fig:InputAndFramework} shows that this speed wave is input into a GCN architecture, which includes two layers of attention mechanisms and spatio-temporal convolution operations. The information transmission between spatial dependence and temporal dependence is fed to a bi-directional RNN to  deal with both forward and backward dependencies in the time sequence for each node.

\subsection{Road Network as Graph}

A road network is represented as a graph, $G = (V,E)$, where the vertex set $V$ contains the $N$ road segments, and the edge set $E$ contains the connecting road segments. The graph is weighted, with the weights encoding the distances between edge-connected vertices. Specifically, in the weighted adjacency matrix $A = (A_{v_iv_j}, \cdots, A_{v_n v_n}) \in R^{N \times N}$, $A_{v_iv_j}$ encodes the spatial correlation (measured by road distance) between vertex $v_i$ and vertex $v_j$. 

\subsection{Integrating Information via Feature Maps}

A novel methodological component of our work is that we leverage the notion of a feature map $X$ so as to account for generalized information, such as traffic speed, traffic flow, the number of lane closures, weather, or, potentially, more. The feature map $X$ correlates with the defined graph structure; that is, $X = (X_{v_it_j}, \cdots, X_{v_n t_k}) \in \mathcal{R}^{N \times T}$, where $N$ is the number of road segments as above, $T$ is the length of the time series, and $X_{v_it_j}$ represents the value of a feature at the $i^{\rm{th}}$ vertex at the $j^{\rm{th}}$ time step. 

\subsubsection{Speed Feature Map $X^S$} For traffic forecasting, the speed $X^S_{v_i,t_j}$ of traffic at vertex $v_i$ during time $t_j$ is assumed to be related to the speeds $X^S_{v,t_1, \dots, t_{j-1}}$ of all road segments $v \in V$ for prior time steps $t_1$ to $t_{j-1}$. We utilize a sliding windows approach to generate the input feature maps, as illustrated on the top panel of Figure~\ref{fig:InputAndFramework}. As the illustration shows, values are obtained at intervals/time slices of $5$ minutes that we refer to as the time step size. We introduce two hyperparameters, $H$ and $P$, to indicate the length of the time series for training and prediction, respectively, in terms of number of time steps.

\subsubsection{Construction Feature Map $X^C$} A similar approach is used to model the construction feature map $X^C$; the presence of a downtime event was initially considered as a binary feature; however, a purely binary 0/1 feature creates numerical problems due to the sparsity of the resulting map. To overcome this, we define the construction work feature map as:
\begin{equation}
	X^{c}_{v} = \max(0,1 - (\frac{dis(v_i,v_j)}{\lambda})^2),
\end{equation} where $\lambda$ is a hyperparameter that weights the relevance of construction at a given nodal distance dis($v_i$,$v_j$), where dis($v_i$,$v_j$) defines the geometric distance between nodes $v_i$ and $v_j$.

\subsubsection{Feature Fusion} The feature maps corresponding to speed and construction work are continuous values. So, we design a feature fusion function to measure the weight ratio of each feature map, defined as: $\hat{X_s} = W_s \odot X^s + W_c \odot X^c$, where $\hat{X_s}$ denotes speed function, $\odot$ is the Hadamard product, and $W_s$ and $W_c$ are learning parameters reflecting the influence degrees of maintenance downtime on the forecasting traffic states. We consider different fusion functions and relate a corresponding ablation study in Section~\ref{sec:Experiments}.

\subsection{Multi-Head Spatial-Temporal Attention Mechanism in GCN-RWZ} 

GCN-RWZ uses a combined spatial and temporal attention mechanism known as "multi-head" attention~\cite{vaswani2017attention,luong2015effective, gu2018recent}. The mechanism is very popular for NLP, as the transformer architecture allows encoding multiple relationships among the input data.  Multi-head attention allows models to learn information in different subspaces, it helps our model to learn spatial relevance in more complex transportation corridor structures. An additional advantage of multi-head attention is that it is relatively computationally efficient.

In essence, the spatial-temporal attention mechanism allows the neural network to pay more attention to the more valuable information. The input, adjusted by the attention mechanism, is fed into the spatial-temporal convolution operations. Graph convolution operates over the spatial dimension, so as to capture spatial dependencies, and temporal convolution operates over the temporal dimension to capture temporal dependencies. We describe each of these operators next.

\subsection{Spatial Convolution} 

To learn the topological relationships in a traffic network, graph convolutional operations are performed on the input feature map from the training data. One first designs a Laplacian matrix of the graph to derive the Laplacian operator and then perform eigendecomposition by Fourier transform. The Laplacian matrix $L_{n\times n}$ is defined as $L = D - W$, where $D$ is the degree matrix, and $W$ is the adjacency matrix of the graph. Specifically:
\vspace*{-2mm}
\begin{equation}
L = D - A = U \Lambda U^{-1} = U\begin{bmatrix}
\lambda_n\\
&\ddots&\\
&&\lambda_n\\
\end{bmatrix}U^{-1}, 
\end{equation}
where $U \in R^{N \times N}$ is the matrix of eigenvectors ordered by eigenvalues, and $\Lambda$ is the diagonal matrix of eigenvalues. 

Let us represent the signal over the graph $G$ at time $t$ as $x = \bf{x}_t^f \in \mathcal{R}^N$. The graph Fourier transform of the signal is then $\hat{x} = U^T x$. Since $U$ is an orthogonal matrix, the corresponding inverse Fourier transform is $x = U\hat{x}$. Based on this, the signal $x$ on the graph
$G$ is filtered by a kernel $g \in R^N$, and the graph convolutions are defined as:
\vspace*{-2mm}
\begin{equation}
x \ast g = f^{-1}(f(x)\odot f(g)) = U(U^{\mathsf{T}} x \odot U^{\mathsf{T}} g),
\end{equation} 
where $\ast$ is graph convolution operation and $\odot$ is the Hadamard product. If we define $U^{\mathsf{T}} g$ as $g_{\theta}$, which is a learnable convolution kernel, the graph convolution is written as: 
\vspace*{-2mm}
\begin{equation}
(x \ast g)_G = U g_{\theta} U^{\mathsf{T}} x.
\end{equation}

Although the simple operation is theoretically feasible, the computational cost is high, because each sample needs feature decomposition, and each forward propagation needs to calculate the product of $U$, $g_{\theta}$, and $U^{\mathsf{T}}$. Inspired by work in~\cite{defferrard2016convolutional, hammond2011wavelets}, the $g_{\theta}$ can be expanded by Chebyshev polynomials, which is defined as:
\vspace*{-2mm}
\begin{equation}
    g_{\theta}(\Lambda)\approx \sum_{k=0}^{K} \theta_k T_K (\widehat{\Lambda}),
\end{equation}
where $\widehat{\Lambda} = \frac{2\Lambda}{\lambda_{max}} - I_N$, $\lambda_{max}$ is the spectral radius, $\theta$ is the vector of Chebyshev coefficient, $T_K$ is defined as $T_k(x) = 2xT_{k-1} - T_{k-2}(x)$, where $T_0(x) = 1$ and $T_1(x) = x$. 
Thus, the graph convolution operation is denoted as: 
\vspace*{-2mm}
\begin{equation}
    (x \ast g)_G = \sum_{k=0}^{K} \theta_k T_K (\widehat{L})x,
\end{equation}
where $\widehat{L} = \frac{2L}{\lambda_{max}} - I_N = U\widehat{\Lambda}U^{\mathsf{T}}$. 

To further improve the computational efficiency based on the Chebyshev model, we use a layer-wise linear model as in~\cite{kipf2016semi}. When $K$ = 1 and $\lambda_{max}$ = 2, the operation is written as: 
\vspace*{-2mm}
\begin{equation}
    (x \ast g)_G = \theta_0 x - \theta_1 D^{-\frac{1}{2}}AD^{-\frac{1}{2}}x
\end{equation}. 

Subsequently, we set $\theta = \theta_0 = -\theta_1$ to avoid overfitting and use a renormalization trick to avoid gradient explosion and disappearance. The operation is then: 
\vspace*{-2mm}
\begin{equation}
    I_N + D^{-\frac{1}{2}}AD^{-\frac{1}{2}} = \widetilde{D}^{-\frac{1}{2}}\widetilde{A}\widetilde{D}^{-\frac{1}{2}},
\end{equation}
where $\widetilde{A} = A + I_N$, $\widetilde{D}_{ii} = \sum_{j}\widetilde{A}_{ij}$. 

In summary, the graph convolutional operations in our model are: 
\begin{equation}
H^{(l+1)} = f(H^l,A) = \sigma ((x \ast g)_G) = \sigma(\widetilde{D}^{-\frac{1}{2}}\widetilde{A}\widetilde{D}^{-\frac{1}{2}}H^l\theta^l)
\end{equation}
where $H^l$ is the output of layer $l$ , $\theta$ is a learnable weight, and $\sigma$ is the sigmoid function.

\subsection{Temporal Convolution}

After spatial convolution is carried out, temporal dependencies are computed through a standard convolutional operation per \cite{zhao2019t}. The output after the spatial-temporal convolution is written as: 
\vspace*{-2mm}
\begin{equation}
\Bar{X}^{(l+1)}_H = \sigma(\Phi \ast (\sigma((x \ast g)_G X^{(l)}_H))) \in R^{C \times N \times T},
\end{equation} 
where $\Phi$ is a parameter of the temporal convolution kernel, the first $\ast$ is a standard convolution operation, $(x \ast g)_G$ represents the graph convolution operation, and $\sigma$ is the ReLU activation function. 

\subsection{Residual Work} 

After all convolutional operations are completed, the resulting output $\Bar{X}^{(l+1)}_H$ of the convolution is fed back through an additional layer of spatial and temporal convolutions. A 1x1 convolutional layer is then used to reduce the output channel to a single dimension. Subsequently, a bidirectional recurrent neural network is used to learn the dynamic behavior in the time sequence for each node. A final linearization function ensures that the output has the same dimension as the prediction. The Adam algorithm is chosen for training, with a learning rate of $0.001$, determined empirically.
\section{Experiments}
\label{sec:Experiments}

\subsection{Datasets}

We utilize the following three datasets. The Tyson's dataset is a new dataset that we have compiled. It is aggregated from 01/01/2019 to 12/31/2019, collected at 5-minute intervals over $131$ mile-segments in the Tyson's Corner region of Northern Virginia. This dataset allows for traffic flow predictions at $15$-, $30$-, and $60$-minute intervals into the future (which, as we further describe below correspond to $3$, $6$, and $12$ time steps, respectively, in our GCN-RWZ model). We process this dataset to remove instances where traffic accidents occurred simultaneously with construction. Min-Max normalization is also applied to control for data imbalance.

The other two datasets do not include roadwork zones and so only allow traffic speed prediction, however they permit us to comparatively evaluate our model against other state-of-art approaches. Specifically, the Los Angeles dataset (Los-loop), debuted in~\cite{zhao2019t} to evaluate the T-GCN model, contains $207$ segments with $5$-minute traffic speed recordings from 03/01/2012 to 03/07/2012. The third dataset, PEMS-BAY, is collected by California Transportation Agencies Performance Measurement System(PeMS) and is a benchmark dataset for many models. The dataset is collected over $325$ segments in the Bay Area and contains $5$-minute recordings from 01/01/2017 to 05/31/2017. We note that we considered a fourth dataset, METR, as a possible benchmark dataset, but METR contains no recordings for many segments at different times. 

On each of these three datasets, $70$\% of the data is used for model training, $10$\% for validation, and $20$\% for testing. 

\subsection{Performance Metrics}

Traffic speed prediction performance was quantified using three metrics: Root Mean Squared Error (RMSE), Mean Absolute Error (MAE), and Mean Absolute Percentage Error (MAPE). Note that since our prediction is traffic speed, the unit for RMSE and MAE, which measure the error between predicted and ground-truth/true speed in a network segment, is miles per hour (MPH). In contrast, MAPE considers not only the error between the predicted and true speed, but also the ratio of the error to the true value. The smaller MAPE is, the better the prediction performance of a model. 

\subsection{Evaluation Setting}

We compare GCN-RWZ to $4$ SOTA models: T-GCN~\cite{zhao2019t}, STGCN~\cite{yu2017spatio}, GraphWaveNet~\cite{wu2019graph}, and ASTGCN~\cite{guo2019attention}. We note that none of these other models account for construction work zones. As we describe in Section~\ref{sec:Methods}, our model, GCN-RWZ, is notably different from these other four SOTA models, particularly in its use of multi-head attention and the configuration of the convolutional operators. T-GCN uses a GRU to learn dynamic traffic flow; STGCN does not include an attention mechanism. GraphWaveNet uses an adaptive dependency matrix and node embedding to capture the hidden spatial dependency in the data and then feeds the information to a dilated graph convolution; ASTGCN uses Chebyshev polynomials and a standard convolution neural network and was the original basis for the GCN-RWZ architecture. In our comparative evaluation, we include a sixth model, to which we refer as GCN-RWZ$^{-}$. This model is a variant of GCN-RWZ that does not include a feature map of work zone data. We include it in our comparative evaluation below to illustrate the behavior of the model in circumstances where construction data is not available, such as for the Los-loop and PEMS-BAY datasets.

\subsection{Performance Comparison}

\begin{figure*}[htbp]
\centering
  \begin{tabular}{ccc}  
  \textbf{Tyson's Corner} & \textbf{Los-loop} & \textbf{PEMS-BAY}\\[-1mm]
  \includegraphics[width=0.33\textwidth]{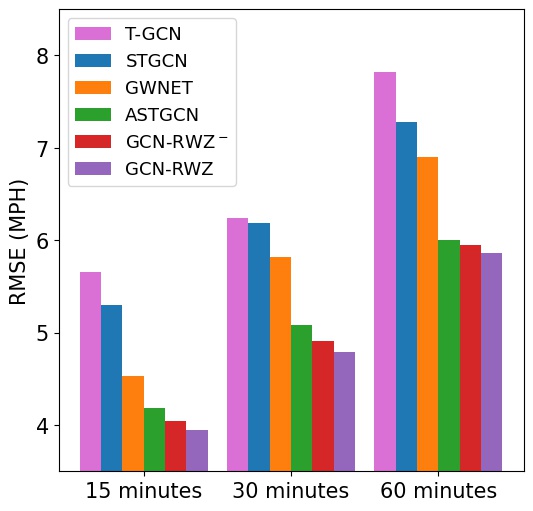} &
\hspace*{-5mm}
  \includegraphics[width=0.33\textwidth]{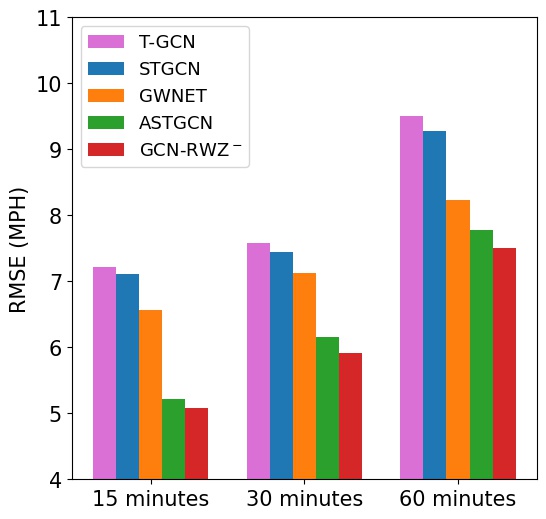} &
\hspace*{-5mm}
\includegraphics[width=0.33\textwidth]{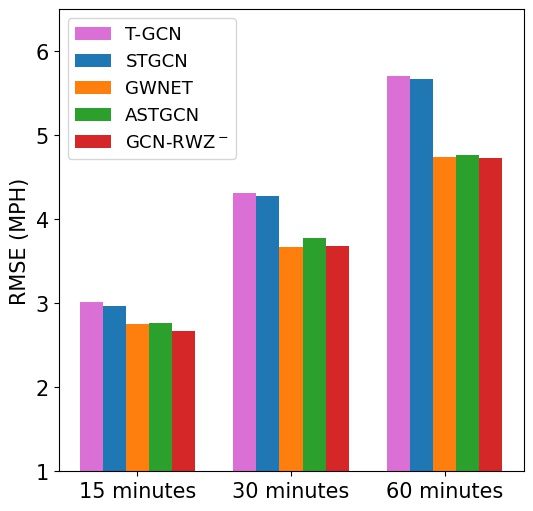}\\[-3mm]
  \end{tabular}
\caption{RMSE comparison for three forecast lengths.}
  \label{RMSE_bar}
\end{figure*}

We relate our comparative evaluation in Table~\ref{tab:freq} and then graphically in Figure~\ref{RMSE_bar}. Table~\ref{tab:freq} relates MAE, (Mean) RMSE, and MAPE over the testing dataset for Tyson's Corner, Los-Loop, and PEMS-BAY for three forecast settings, $15$-, $30$-, and $60$-minutes. Figure~\ref{RMSE_bar} provides a closer look and shows the RMSE over all models. We emphasize that all RMSE values reported, including for the other SOTA models, are mean RMSEs; in some work, such as in~\cite{zhao2019t}, the authors occasionally report the minimum RMSE when relating the performance of T-GCN, but such a value always provides a rosier view of performance. To additionally be consistent with other works that report the performance of the other models, we utilize mean RMSE.

\begin{table*}[htbp]
    \renewcommand\arraystretch{1.3}
    \small
  \centering
  \setlength\tabcolsep{5pt}
  \caption{Performance Comparison, Tyson's Corner Dataset and Los-loop Dataset. Best value per metric is highlighted in boldface font.}
  \label{tab:freq}
  \begin{tabular}{|p{3em}|p{6.2em}|p{2em}p{2em}p{4.9em}|p{2em}p{2em}p{4.9em}|p{2em}p{2em}p{4.9em}|}
    \hline
    Dataset & Model 
    & \multicolumn{3}{c|}{15-min forecast}
    & \multicolumn{3}{c|}{30-min forecast}
    & \multicolumn{3}{c|}{60-min forecast} \\
    \cline{3-11}
      & & MAE& RMSE & MAPE  (\%) & MAE& RMSE & MAPE (\%) & MAE& RMSE & MAPE (\%)
    \\ 
    \hline
&T-GCN & 3.42 & 5.66 & 12.21 & 3.62 & 6.24 & 13.57 & 5.33 & 7.82 & 16.30\\
&STGCN & 3.35 & 5.07 & 12.03 & 3.58 & 6.06 & 13.45 & 4.85 & 7.02 & 15.82 \\

TYSON & GraphWaveNet & 2.92 & 4.53 & 10.10 & 3.52 & 5.82 & 12.32 & 4.55 & 6.90 & 14.97 \\
&ASTGCN  & 2.68 & 4.18 & 09.50 & 3.18 & 5.08 & 11.58 & 3.73 & 6.00 & 13.34 \\
&GCN-RWZ$^-$  & 2.62 & 4.04 & 08.83 & 3.08 & 4.91 & 10.67 & 3.70 & 5.95 & 13.30 \\
&GCN-RWZ  & \textbf{2.56} & \textbf{3.95} & \textbf{08.68} & \textbf{3.01} & \textbf{4.79} & \textbf{10.33} & \textbf{3.49} & \textbf{5.86} & \textbf{12.82}\\
    \hline
& T-GCN & 4.58 & 7.22 & 12.68 & 5.35 & 7.58 & 14.38 &  5.52 & 9.50 & 15.04\\
& STGCN& 4.52 & 7.11 & 12.51 & 4.65 & 7.44 & 13.81 & 5.38 & 9.28 & 14.73\\

LOS & GraphWaveNet & 3.40 & 6.57 & 10.12 & 4.43 & 7.13 & 12.55 & 4.81 & 8.23 & 14.29 \\
& ASTGCN & 3.08 & 5.22 & 08.51 & 3.59 & 6.15 & 10.25 & 4.56 & 7.78 & 13.13 \\
& GCN-RWZ$^-$ & \textbf{2.92} & \textbf{5.08} & \textbf{07.77} & \textbf{3.39} & \textbf{5.92} & \textbf{09.05} & \textbf{4.49} & \textbf{7.51} & \textbf{12.79}\\
    \hline

& T-GCN & 1.45 & 3.02 & 03.14 & 1.88 & 4.31 & 04.21 & 2.50 & 5.71 & 05.79\\
& STGCN& 1.36& 2.97 & 02.91 & 1.82 & 4.28 & 04.16 & 2.48 & 5.67 & 05.75\\
PEMS & GraphWaveNet & 1.30 & 2.75 & 02.73 & 1.66 & 3.70 & 03.67 & 2.11 & 4.74 & 04.92 \\
& ASTGCN & 1.31 & 2.76 & 02.75 & 1.75 & 3.78 & 03.85 & 2.13 & 4.76 & 05.15 \\
& GCN-RWZ$^-$ & \textbf{1.28} & \textbf{2.67} & \textbf{02.63} & \textbf{1.58} & \textbf{3.68} & \textbf{03.54} & \textbf{2.09}& \textbf{4.73}  & \textbf{04.90}\\
    \hline
\end{tabular}
\end{table*}

Table~\ref{tab:freq} and Figure~\ref{RMSE_bar} support the following observations. First, the performance of all models deteriorates with longer forecast length. This is not surprising. The nonlinear characteristics of the model become more complicated as the forecast length increases. Second, as we highlight with boldface font the lowest value for a metric in Table~\ref{tab:freq}, the best-performing model across all metrics and forecast lengths on the Tyson's dataset is our model, GCN-RWZ.  The difference in performance is significant with longer forecast length. This is also not surprising and is due to the attention mechanism in GCN-RWZ, which focuses limited attention on key information and so helps the model obtain effective information quickly. In addition, it is helpful that  each subsequent step of the calculation does not rely on the previous calculations, which is another benefit of the multi-head attention mechanism. On the other two datasets, we can only compare  GCN-RWZ$^{-}$  to the other models. Table~\ref{tab:freq} and Figure~\ref{RMSE_bar} show that GCN-RWZ$^{-}$  outperforms the other models on the Los-Loop dataset on all forecast lengths. On the PEMS-BAY dataset, many other models come close in performance to GCN-RWZ$^{-}$. This is due to several reasons. Since the time of recording traffic speed in the PEMS-BAY dataset is as long as six months, most traffic speeds are less affected by anomalous conditions. As the number of traffic sensors increases, GraphWaveNet model can handle very long sequences due to its receptive fields and adaptive dependency matrix. Thus, its performance is better than ASTGCN and is similar to our model, GCN-RWZ$^{-}$.
Altogether, the results shown in Table~\ref{tab:freq} and Figure~\ref{RMSE_bar} allow us to conclude that both the GCN-RWZ and GCN-RWZ$^{-}$ have superior performance over the other models, and, in particular, as observed over the Tyson’s Corner dataset, there is a clear and measurable improvement when construction work zone data are included in the model. 

\begin{figure*}[htbp]
  \centering
  \begin{tabular}{cc}  
  \includegraphics[width=\columnwidth]{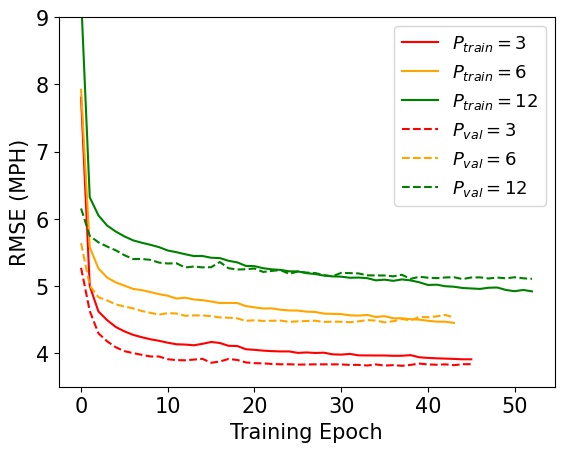} &
  \includegraphics[width=\columnwidth]{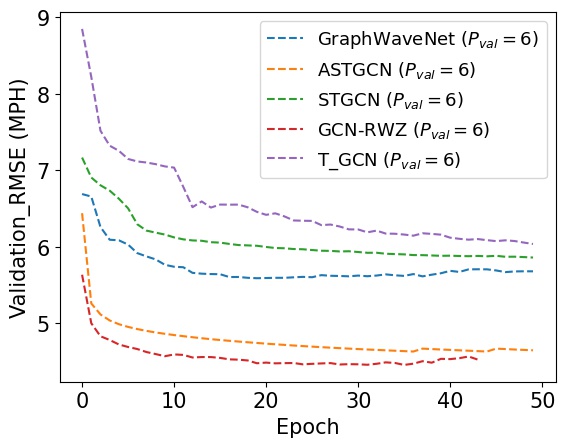}\\[-3mm]
  \end{tabular}
   \caption{Left: RMSE over the training epochs for GCN-RWZ, separating the curves for the different forecast lengths (15, 30, and 60 minutes), and doing so for both the training (train) and validation (val) dataset; P refers to the forecast length; a value of $3$ corresponds to $3$ time steps of size $5$ minutes each and so to a forecast length of $15$ minutes. Right: Comparison of RMSE over the validation dataset over the training epochs for a forecast length of $30$ minutes over the various models.}
  \label{RMSE}
\end{figure*}

Figure~\ref{RMSE} provides further information. First, the left panel  shows the reduction in RMSE over the training epochs for GCN-RWZ, separating the different forecast lengths, and doing so for both the training and validation dataset. The results relate that the model converges to low RMSEs; as expected lower RMSEs are obtained on the shorter forecast lengths. The right panel of Figure~\ref{RMSE} compares the various models, showing the respective RMSEs on the validation dataset over the training epochs for the forecast length of $30$ minutes. The results support the comparative analysis related above; they show that GCN-RWZ and ASTGCN achieve the lowest RMSEs on the validation dataset over all the other models, with GCN-RWZ outperforming ASTGCN.

\subsection{Detailed Evaluation of GCN-RWZ}
 
\begin{figure*}[htbp]
  \centering
  \begin{tabular}{c}
  \includegraphics[width=0.5\textwidth]{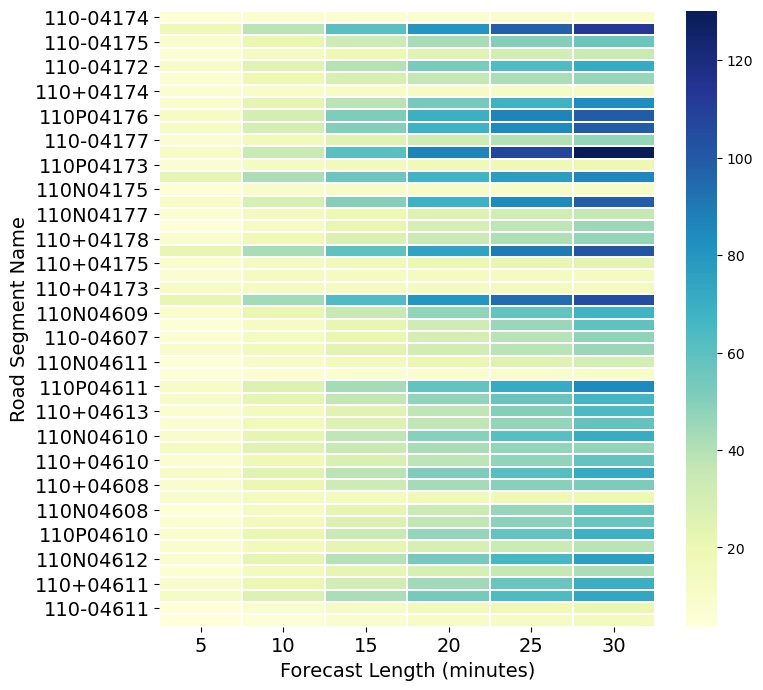}\\[-3mm]
  \end{tabular}
  \caption{Heatmap representation of GCN-RWZ RMSE per segment and time of day (x axis) over increasing forecast lengths (y axis), Tyson's Corner dataset}
  \label{heatmap}
\end{figure*}

We now focus on the performance of GCN-RWZ on individual segments of the Tyson’s dataset and relate this performance via a heatmap in Figure~\ref{heatmap}. The heatmap shows the RMSE per segment (labeled on y axis to indicate both road and time of day) over increasing forecast lengths (x axis).  Figure~\ref{heatmap} shows that the highest errors occur on highway road segments. For example, the results obtained for the road segments "110+04174" and "110P04177" on I-66 are worse than those obtained for the road segment "110+05695,", which is located on State Route 7. The performance on segments on I-495 is better than on segments on I-66. 

\begin{figure}[htbp]
\centering
\includegraphics[width=\columnwidth]{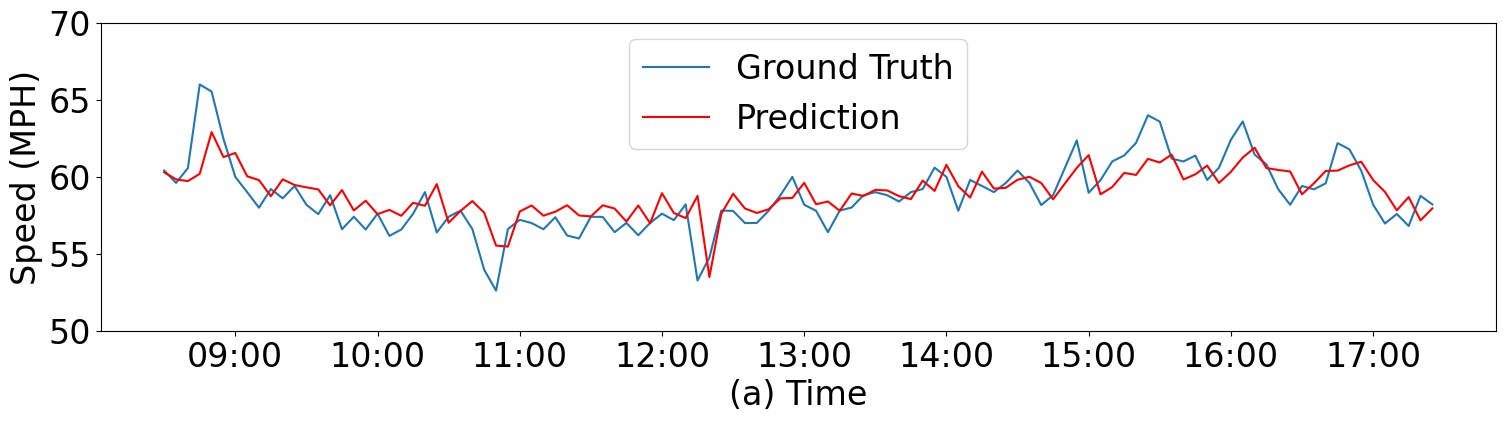}\\
\includegraphics[width=\columnwidth]{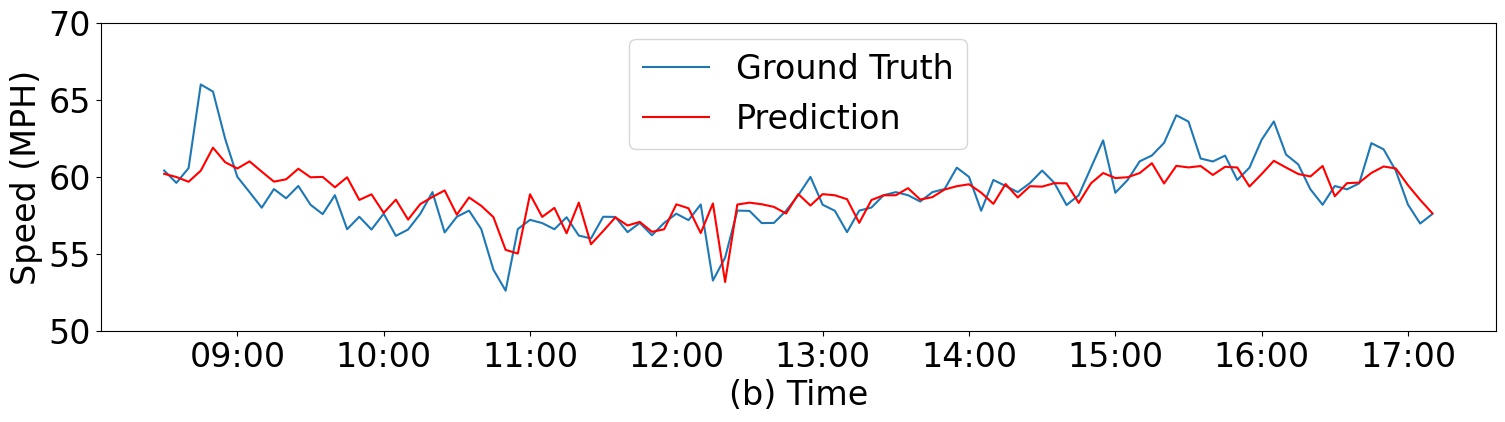} \\
\includegraphics[width=\columnwidth]{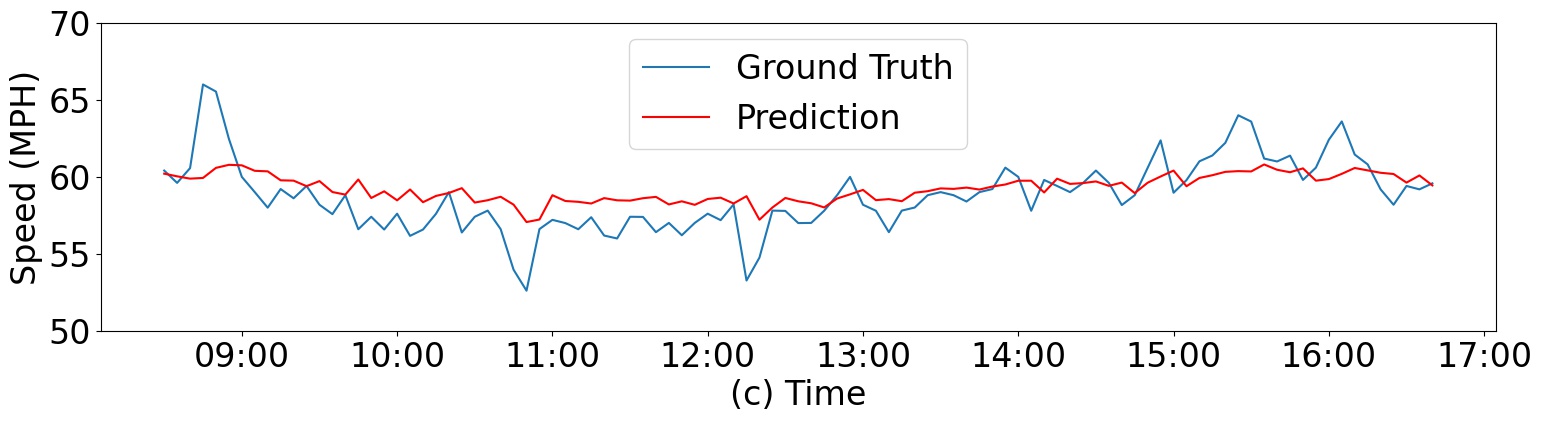}\\[-3mm]
\caption{Relative Speed Forecasting Accuracy on Road Segments "110+04174" on 11/22/2019: (a) 15 minute forecasts (b) 30 minute forecasts (c) 60 minute forecasts}
\label{speed_info_tmc2}
\end{figure}

\begin{figure}[htbp]
\centering
\includegraphics[width=\columnwidth]{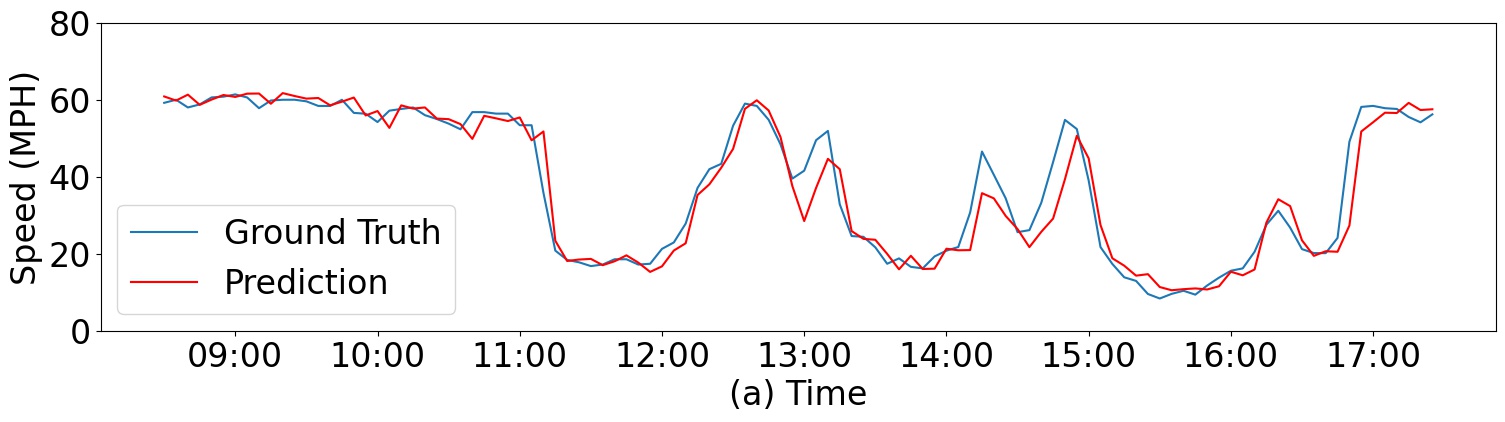}\\
\includegraphics[width=\columnwidth]{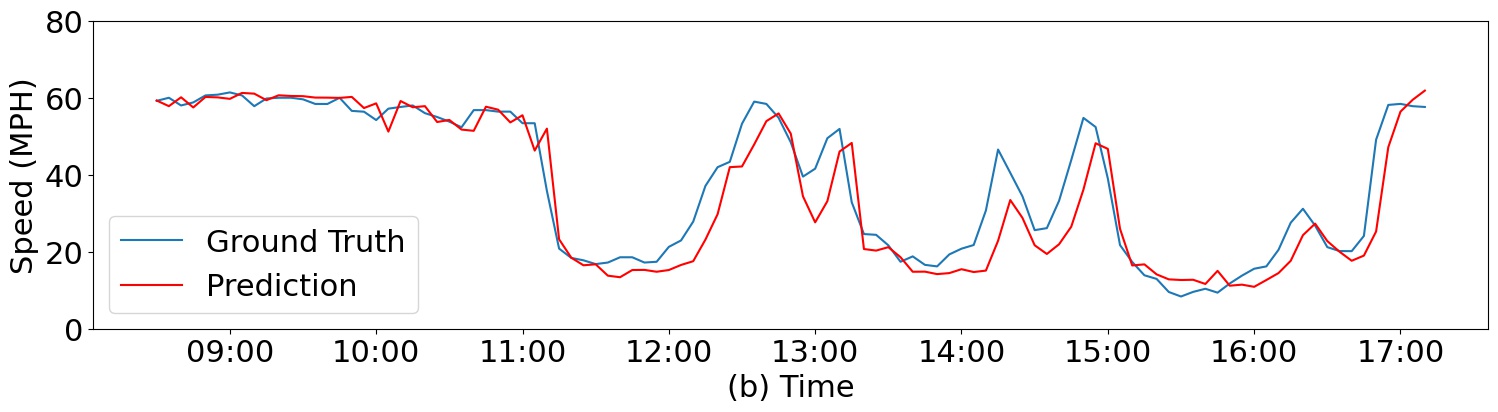} \\
\includegraphics[width=\columnwidth]{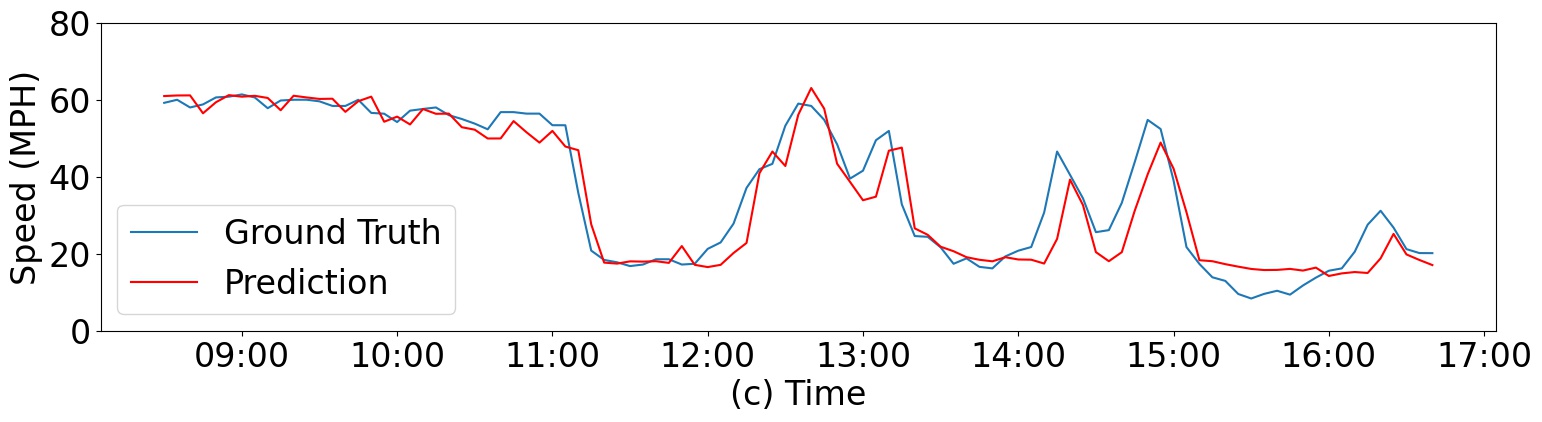}\\[-3mm]
\caption{Relative speed forecasting accuracy on road segment 110P04611 on 11/22/2019: (a) 15 minute forecasts (b) 30 minute forecasts (c) 60 minute forecasts.}
\label{speed_info_tmc3}
\end{figure}

Figures~\ref{speed_info_tmc2} and \ref{speed_info_tmc3} show in detail how  GCN-RWZ performs on forecasting lengths of 15 minutes, 30 minutes, and 60 minutes, by superimposing the ground truth with the model-predicted speed on different road segments. As expected, the prediction accuracy is best for the shorter forecast of 15 minutes; however, accuracy is maintained at a relatively high level even for the longer forecasts of 30 and 60 minutes.

\subsection{Ablation Study}

One of the challenges in modern deep learning is the number of algorithmic and modeling decisions that must be made. An ablation study is undertaken to understand the sensitivity of the GCN-RWZ model to these decisions. Two analyses are considered regarding construction work zone feature map characterization and the speed wave fusion method. The characterization of the speed wave function determines the ability of the model to learn in a specific environment. Table~\ref{speedwave} shows the performance of the model using three different functions for feature fusion. The best RMSE, MAE, and MAPE are obtained on the first setting of a learnable weight matrix for each feature map. 

\begin{table}[htbp]
  \renewcommand\arraystretch{1.3}
   \small
  \centering
  \setlength\tabcolsep{11pt}
\caption{ Ablation study on values for three speed wave $\hat{X_s}$ function}
\vspace{1mm}
\label{speedwave}
 \begin{tabular}{|p{8em}|p{2em}|p{2em}|p{5em}|} 
 \hline
 Speed Wave & RMSE & MAE & MAPE (\%) \\ [0.3ex] 
 \hline
$W_s \odot X^s + W_c \odot X^c$ & \textbf{4.79} & \textbf{3.01} & \textbf{10.33} \\ 
 \hline
$X^s + W_c \odot X^c$& 4.93 & 3.11 & 10.41 \\
 \hline
$X^s \odot X^s + W_c $& 4.98 & 3.17 & 10.53 \\
 \hline
\end{tabular}
\end{table}

Of particular note is the characterization of the workzone feature map. The feature map parameterization function requires a control hyperparameter, $\lambda$, which reflects the impact of construction workzones on nearby road segments. Table~\ref{lamda} shows the performance of the model using $\lambda$ = 1, 3, 5, 7 for the 30-minute forward forecast. While performance differences are small, the smaller values of $\lambda$ lead to improved performance while still allowing us to address the numerical issues associated with a purely binary feature map. 

\begin{table}[htbp]
  \renewcommand\arraystretch{1.3}
  \small
  \centering
  \setlength\tabcolsep{15pt}

\caption{Ablation Study on Values for Feature Map Parameter $\lambda$ }
\label{lamda}
\vspace{1mm}
 \begin{tabular}{|p{1em}|p{2em}|p{2em}|p{5.5em}|} 
 \hline
 $\lambda$ & RMSE & MAE & MAPE (\% ) \\ [0.3ex] 
 \hline
 $1$ & 4.91 & 3.13 & 10.54\\ 
 \hline
 $3$ & \textbf{4.79} & \textbf{3.01} & \textbf{10.33} \\ 
 \hline
 $5$ & 4.96 & 3.18 & 10.62 \\
 \hline
 $7$ & 5.05 & 3.22 & 11.08 \\
 \hline
\end{tabular}
\end{table}

These studies indicate that the modeling decisions delineated in Section~\ref{sec:Methods} (for instance, the form of the convolutional polynomial) lead to predictive modeling improvements. The GCN-RWZ model presented in this report is the result of the above ablation study.

\section{Conclusions}
\label{sec:Conclusions}

We have proposed here GCN-RWZ, a GCN-based model that captures the complex spatio-temporal relationships of large-scale traffic networks. GCN-RWZ leverages a graph structure representation of traffic flow across a transportation corridor. In contrast to existing SOTA methods, the graph representation of traffic flow (speed) in GCN-RWZ is fused with a graph model of a construction workzone on any segment within the network. The fusion results in a time-history “speed wave” that serves as input to the GCN-RWZ learning algorithm. The GCN-RWZ architecture includes many refinements and advancements over existing approaches and is the first such model designed to incorporate workzone impact information in a flexible and generalizable fashion. 

The GCN-RWZ model was tested on three representative datasets and compared against several SOTA models that serve as benchmarks for traffic flow prediction. The GCN-RWZ showed measurably better performance in traffic speed prediction compared to any of the benchmark models on all datasets. This improvement existed even when construction workzone data was not incorporated into the model. When workzone data was included, the resulting improvement was significant and consistent. This suggests that the developed model is a viable platform for further studies, refinements, and implementation. It also suggests that adding additional traffic network information through fused feature maps is a valuable approach to improving predictive performance.


\IEEEpeerreviewmaketitle


\section*{Acknowledgment}
This work is supported in part from a grant to DL from the Virginia Transportation Research Council (VTRC). The authors would like to thank Michael Fitch and Michael Fontaine of VTRC for their guidance. This material is additionally based upon work by AS supported by (while serving at) the National Science Foundation. Any opinion, findings, and conclusions or recommendations expressed in this material are those of the author(s) and do not necessarily reflect the views of the National Science Foundation. 

\ifCLASSOPTIONcaptionsoff
  \newpage
\fi

\bibliographystyle{IEEEtran}
\bibliography{ref.bib}

\begin{thebibliography}{10}
\providecommand{\url}[1]{#1}
\csname url@samestyle\endcsname
\providecommand{\newblock}{\relax}
\providecommand{\bibinfo}[2]{#2}
\providecommand{\BIBentrySTDinterwordspacing}{\spaceskip=0pt\relax}
\providecommand{\BIBentryALTinterwordstretchfactor}{4}
\providecommand{\BIBentryALTinterwordspacing}{\spaceskip=\fontdimen2\font plus
\BIBentryALTinterwordstretchfactor\fontdimen3\font minus
  \fontdimen4\font\relax}
\providecommand{\BIBforeignlanguage}[2]{{%
\expandafter\ifx\csname l@#1\endcsname\relax
\typeout{** WARNING: IEEEtran.bst: No hyphenation pattern has been}%
\typeout{** loaded for the language `#1'. Using the pattern for}%
\typeout{** the default language instead.}%
\else
\language=\csname l@#1\endcsname
\fi
#2}}
\providecommand{\BIBdecl}{\relax}
\BIBdecl

\bibitem{Zhang11}
``Data-driven intelligent transportation systems: A survey,'' vol.~12, no.~4,
  p. 1624–1639, 2011.

\bibitem{cui2019traffic}
Z.~Cui, K.~Henrickson, R.~Ke, and Y.~Wang, ``Traffic graph convolutional
  recurrent neural network: A deep learning framework for network-scale traffic
  learning and forecasting,'' \emph{IEEE Transactions on Intelligent
  Transportation Systems}, 2019.

\bibitem{li2017diffusion}
Y.~Li, R.~Yu, C.~Shahabi, and Y.~Liu, ``Diffusion convolutional recurrent
  neural network: Data-driven traffic forecasting,'' \emph{arXiv preprint
  arXiv:1707.01926}, 2017.

\bibitem{yu2017spatio}
B.~Yu, H.~Yin, and Z.~Zhu, ``Spatio-temporal graph convolutional networks: A
  deep learning framework for traffic forecasting,'' \emph{arXiv preprint
  arXiv:1709.04875}, 2017.

\bibitem{diao2019dynamic}
Z.~Diao, X.~Wang, D.~Zhang, Y.~Liu, K.~Xie, and S.~He, ``Dynamic
  spatial-temporal graph convolutional neural networks for traffic
  forecasting,'' in \emph{Proceedings of the AAAI Conference on Artificial
  Intelligence}, vol.~33, 2019, pp. 890--897.

\bibitem{guo2019attention}
S.~Guo, Y.~Lin, N.~Feng, C.~Song, and H.~Wan, ``Attention based
  spatial-temporal graph convolutional networks for traffic flow forecasting,''
  in \emph{Proceedings of the AAAI Conference on Artificial Intelligence},
  vol.~33, 2019, pp. 922--929.

\bibitem{schrank_2019_2019}
D.~Schrank, B.~Eisele, and T.~Lomax, ``2019 urban mobility report,'' p.~50,
  2019.

\bibitem{du2017predicting}
B.~Du, S.~Chien, J.~Lee, and L.~Spasovic, ``Predicting freeway work zone delays
  and costs with a hybrid machine-learning model,'' \emph{Journal of Advanced
  Transportation}, vol. 2017, 2017.

\bibitem{tong2008highway}
M.~Tong and H.~Xue, ``Highway traffic volume forecasting based on seasonal
  arima model,'' \emph{Journal of Highway and Transportation Research and
  Development (English Edition)}, vol.~3, no.~2, pp. 109--112, 2008.

\bibitem{wu2004travel}
C.-H. Wu, J.-M. Ho, and D.-T. Lee, ``Travel-time prediction with support vector
  regression,'' \emph{IEEE transactions on intelligent transportation systems},
  vol.~5, no.~4, pp. 276--281, 2004.

\bibitem{moretti2015urban}
F.~Moretti, S.~Pizzuti, S.~Panzieri, and M.~Annunziato, ``Urban traffic flow
  forecasting through statistical and neural network bagging ensemble hybrid
  modeling,'' \emph{Neurocomputing}, vol. 167, pp. 3--7, 2015.

\bibitem{kelejian1999generalized}
H.~H. Kelejian and I.~R. Prucha, ``A generalized moments estimator for the
  autoregressive parameter in a spatial model,'' \emph{International economic
  review}, vol.~40, no.~2, pp. 509--533, 1999.

\bibitem{huang2014deep}
W.~Huang, G.~Song, H.~Hong, and K.~Xie, ``Deep architecture for traffic flow
  prediction: deep belief networks with multitask learning,'' \emph{IEEE
  Transactions on Intelligent Transportation Systems}, vol.~15, no.~5, pp.
  2191--2201, 2014.

\bibitem{tian2015predicting}
Y.~Tian and L.~Pan, ``Predicting short-term traffic flow by long short-term
  memory recurrent neural network,'' in \emph{2015 IEEE international
  conference on smart city/SocialCom/SustainCom (SmartCity)}.\hskip 1em plus
  0.5em minus 0.4em\relax IEEE, 2015, pp. 153--158.

\bibitem{ma2015long}
X.~Ma, Z.~Tao, Y.~Wang, H.~Yu, and Y.~Wang, ``Long short-term memory neural
  network for traffic speed prediction using remote microwave sensor data,''
  \emph{Transportation Research Part C: Emerging Technologies}, vol.~54, pp.
  187--197, 2015.

\bibitem{ma2015large}
X.~Ma, H.~Yu, Y.~Wang, and Y.~Wang, ``Large-scale transportation network
  congestion evolution prediction using deep learning theory,'' \emph{PloS
  one}, vol.~10, no.~3, p. e0119044, 2015.

\bibitem{cui2018deep}
Z.~Cui, R.~Ke, Z.~Pu, and Y.~Wang, ``Deep bidirectional and unidirectional lstm
  recurrent neural network for network-wide traffic speed prediction,''
  \emph{arXiv preprint arXiv:1801.02143}, 2018.

\bibitem{ma2017learning}
X.~Ma, Z.~Dai, Z.~He, J.~Ma, Y.~Wang, and Y.~Wang, ``Learning traffic as
  images: a deep convolutional neural network for large-scale transportation
  network speed prediction,'' \emph{Sensors}, vol.~17, no.~4, p. 818, 2017.

\bibitem{jo2018image}
D.~Jo, B.~Yu, H.~Jeon, and K.~Sohn, ``Image-to-image learning to predict
  traffic speeds by considering area-wide spatio-temporal dependencies,''
  \emph{IEEE Transactions on Vehicular Technology}, vol.~68, no.~2, pp.
  1188--1197, 2018.

\bibitem{vaswani2017attention}
A.~Vaswani, N.~Shazeer, N.~Parmar, J.~Uszkoreit, L.~Jones, A.~N. Gomez,
  {\L}.~Kaiser, and I.~Polosukhin, ``Attention is all you need,'' in
  \emph{Advances in neural information processing systems}, 2017, pp.
  5998--6008.

\bibitem{wu2019graph}
Z.~Wu, S.~Pan, G.~Long, J.~Jiang, and C.~Zhang, ``Graph wavenet for deep
  spatial-temporal graph modeling,'' \emph{arXiv preprint arXiv:1906.00121},
  2019.

\bibitem{pan2019urban}
Z.~Pan, Y.~Liang, W.~Wang, Y.~Yu, Y.~Zheng, and J.~Zhang, ``Urban traffic
  prediction from spatio-temporal data using deep meta learning,'' in
  \emph{Proceedings of the 25th ACM SIGKDD International Conference on
  Knowledge Discovery \& Data Mining}, 2019, pp. 1720--1730.

\bibitem{ruiz2020gated}
L.~Ruiz, F.~Gama, and A.~Ribeiro, ``Gated graph recurrent neural networks,''
  \emph{arXiv preprint arXiv:2002.01038}, 2020.

\bibitem{zheng2020gman}
C.~Zheng, X.~Fan, C.~Wang, and J.~Qi, ``Gman: A graph multi-attention network
  for traffic prediction,'' in \emph{Proceedings of the AAAI Conference on
  Artificial Intelligence}, vol.~34, no.~01, 2020, pp. 1234--1241.

\bibitem{keneshloo2019deep}
Y.~Keneshloo, T.~Shi, N.~Ramakrishnan, and C.~K. Reddy, ``Deep reinforcement
  learning for sequence-to-sequence models,'' \emph{IEEE transactions on neural
  networks and learning systems}, vol.~31, no.~7, pp. 2469--2489, 2019.

\bibitem{velivckovic2017graph}
P.~Veli{\v{c}}kovi{\'c}, G.~Cucurull, A.~Casanova, A.~Romero, P.~Lio, and
  Y.~Bengio, ``Graph attention networks,'' \emph{arXiv preprint
  arXiv:1710.10903}, 2017.

\bibitem{perozzi2014deepwalk}
B.~Perozzi, R.~Al-Rfou, and S.~Skiena, ``Deepwalk: Online learning of social
  representations,'' in \emph{Proceedings of the 20th ACM SIGKDD international
  conference on Knowledge discovery and data mining}, 2014, pp. 701--710.

\bibitem{kang2019learning}
Z.~Kang, H.~Xu, J.~Hu, and X.~Pei, ``Learning dynamic graph embedding for
  traffic flow forecasting: A graph self-attentive method,'' in \emph{2019 IEEE
  Intelligent Transportation Systems Conference (ITSC)}.\hskip 1em plus 0.5em
  minus 0.4em\relax IEEE, 2019, pp. 2570--2576.

\bibitem{wu2020comprehensive}
Z.~Wu, S.~Pan, F.~Chen, G.~Long, C.~Zhang, and S.~Y. Philip, ``A comprehensive
  survey on graph neural networks,'' \emph{IEEE Transactions on Neural Networks
  and Learning Systems}, 2020.

\bibitem{zhao2019t}
L.~Zhao, Y.~Song, C.~Zhang, Y.~Liu, P.~Wang, T.~Lin, M.~Deng, and H.~Li,
  ``T-gcn: A temporal graph convolutional network for traffic prediction,''
  \emph{IEEE Transactions on Intelligent Transportation Systems}, 2019.

\bibitem{zhou2020reinforced}
F.~Zhou, Q.~Yang, K.~Zhang, G.~Trajcevski, T.~Zhong, and A.~Khokhar,
  ``Reinforced spatio-temporal attentive graph neural networks for traffic
  forecasting,'' \emph{IEEE Internet of Things Journal}, 2020.

\bibitem{luong2015effective}
M.-T. Luong, H.~Pham, and C.~D. Manning, ``Effective approaches to
  attention-based neural machine translation,'' \emph{arXiv preprint
  arXiv:1508.04025}, 2015.

\bibitem{gu2018recent}
J.~Gu, Z.~Wang, J.~Kuen, L.~Ma, A.~Shahroudy, B.~Shuai, T.~Liu, X.~Wang,
  G.~Wang, J.~Cai \emph{et~al.}, ``Recent advances in convolutional neural
  networks,'' \emph{Pattern Recognition}, vol.~77, pp. 354--377, 2018.

\bibitem{defferrard2016convolutional}
M.~Defferrard, X.~Bresson, and P.~Vandergheynst, ``Convolutional neural
  networks on graphs with fast localized spectral filtering,'' in
  \emph{Advances in neural information processing systems}, 2016, pp.
  3844--3852.

\bibitem{hammond2011wavelets}
D.~K. Hammond, P.~Vandergheynst, and R.~Gribonval, ``Wavelets on graphs via
  spectral graph theory,'' \emph{Applied and Computational Harmonic Analysis},
  vol.~30, no.~2, pp. 129--150, 2011.

\bibitem{kipf2016semi}
T.~N. Kipf and M.~Welling, ``Semi-supervised classification with graph
  convolutional networks,'' \emph{arXiv preprint arXiv:1609.02907}, 2016.

\end{thebibliography}
\end{document}